\documentclass{article}

\pdfoutput=1

    \PassOptionsToPackage{numbers}{natbib}
    \usepackage[preprint]{neurips_2024}

\usepackage{graphicx}
\usepackage[utf8]{inputenc} % allow utf-8 input
\usepackage[T1]{fontenc}    % use 8-bit T1 fonts
\usepackage{hyperref}       % hyperlinks
\usepackage{url}            % simple URL typesetting
\usepackage{booktabs}       % professional-quality tables
\usepackage{amsfonts}       % blackboard math symbols
\usepackage{nicefrac}       % compact symbols for 1/2, etc.
\usepackage{microtype}      % microtypography
\usepackage[dvipsnames]{xcolor}         % colors
\usepackage[table]{xcolor}

\usepackage{subcaption}
\usepackage{wrapfig}
\usepackage{graphicx}
\usepackage{amsmath} 
\usepackage{multirow}
\usepackage{soul}
\usepackage[normalem]{ulem}
\usepackage{tabularx, makecell, booktabs}
\useunder{\uline}{\ul}{}
\usepackage{tcolorbox}
\usepackage{array}
\usepackage{arydshln}
\newcolumntype{C}[1]{>{\centering\arraybackslash}p{#1}}

\title{MSA: Memory Sparse Attention for Efficient End-to-End Memory Model Scaling to 100M Tokens}

\author{
  Yu Chen${}^{1,2}$\thanks{~ Equal contribution.},
  Runkai Chen${}^{1,2,3}$\footnotemark[1],
  Sheng Yi${}^{1,2}$,
  Xinda Zhao${}^{1,2}$,
  Xiaohong Li${}^{1,2}$,
  Jianjin Zhang${}^{1,2}$, \\
  {\bf Jun Sun}${}^{3}$,
  {\bf Chuanrui Hu}${}^{1,2}$,
  {\bf Yunyun Han}${}^{1,2}$,
  {\bf Lidong Bing}${}^{2}$,
  {\bf Yafeng Deng}${}^{1,2}$\thanks{~ Corresponding authors.},
  {\bf Tianqiao Chen}${}^{2}$\footnotemark[2]
  \\
  ${}^{1}$Evermind \\
  ${}^{2}$Shanda Group \\
${}^{3}$Peking University \\
  \vspace{0.1cm}
  {\tt \{yu.chen, runkai.chen, sheng.yi, xinda.zhao, xiaohong.li\}@shanda.com} \\
  {\tt \{jianjin.zhang, chuanrui.hu, hanyunyun, lidong.bing, dengyafeng, ctq\}@shanda.com} \\
  {\tt sunjun@pku.edu.cn} 
}

\begin{document}

\maketitle

\begin{abstract}

Long-term memory is a cornerstone of human intelligence.
Enabling AI to process lifetime-scale information, reaching hundreds of millions of tokens, remains a long-standing pursuit in the field.
Due to the constraints of full-attention architectures, the effective context length of large language models (LLMs) is typically limited to 1M tokens.
Existing explorations, such as hybrid linear attention, fixed-size memory states (e.g., RNNs), and external storage methods like RAG or agent systems, attempt to extend this limit.
However, these approaches often suffer from severe precision degradation and rapidly increasing latency as context length grows, an inability to dynamically modify memory content, or a lack of end-to-end optimization.
These bottlenecks impede complex scenarios like large‑corpus summarization, Digital Twins with stable personas, and long‑history agent reasoning, while limiting memory capacity and slowing inference.
We present Memory Sparse Attention (MSA), an end-to-end trainable, efficient, and massively scalable memory model framework.
Through core innovations including scalable sparse attention architecture and document‑wise RoPE,
MSA achieves linear complexity in both training and inference while maintaining exceptional precision stability, exhibiting less than 9\% degradation when scaling from 16K to 100M tokens.
Furthermore, KV cache compression, combined with Memory Parallel during inference, enables 100M tokens inference on $2\times \text{A}800$ GPUs.
In addition, we propose a Memory Interleaving mechanism that effectively facilitates complex multi‑hop reasoning across scattered memory segments.
MSA significantly surpasses frontier language models, state-of-the-art (SOTA) RAG systems, and leading memory agents in long-context benchmarks.
These results demonstrate that by decoupling memory capacity from reasoning, MSA provides a scalable foundation to endow general-purpose models with intrinsic, lifetime-scale memory.
\end{abstract}

\section{Introduction}

While Large Language Models (LLMs) have demonstrated remarkable proficiency in competitive mathematical reasoning~\cite{gsm8k, Hendrycks2021MeasuringMP}, collaborative programming~\cite{humaneval, swebench}, and role-playing~\cite{rolellm, generative_agents}, they remain confronted by a formidable challenge: long-term, fine-grained memory retention~\cite{lost_in_the_middle, infinitebench}. Scenarios such as comprehending extensive novel series~\cite{longbench,narrativeqa}, maintaining consistent personas in role‑playing, or managing the long‑term history of multi-agent systems~\cite{camel, generative_agents} place stringent demands on the model's memory capacity, specifically its effective context length. Research in cognitive science estimates the functional information capacity of human memory to be on the order of $10^9$ bits~\citep{landauer1986Howmuchdopeopleremember}. Assuming an effective semantic density of $3$--$5$ bits per token, this corresponds to a lifelong capacity of approximately $200$--$300$ million tokens. Consequently, to truly bridge the gap toward human-scale memory and facilitate applications such as Digital Twins, models must effectively process contexts extending into the hundreds of millions of tokens. In stark contrast, contemporary LLMs typically support effective context lengths ranging from 128k to 1M tokens~\cite{llama3, liu2025deepseek,openai2024gpt4o}. Even architectures explicitly designed for long contexts~\cite{qwen2.5,qwen3-coder}, despite undergoing rigorous training pipelines, rarely exceed the 1M token threshold. To bridge this magnitude of disparity, a specialized mechanism tailored for human-scale memory is imperative.

\begin{figure}[h]
    \centering
    \includegraphics[width=1\linewidth]{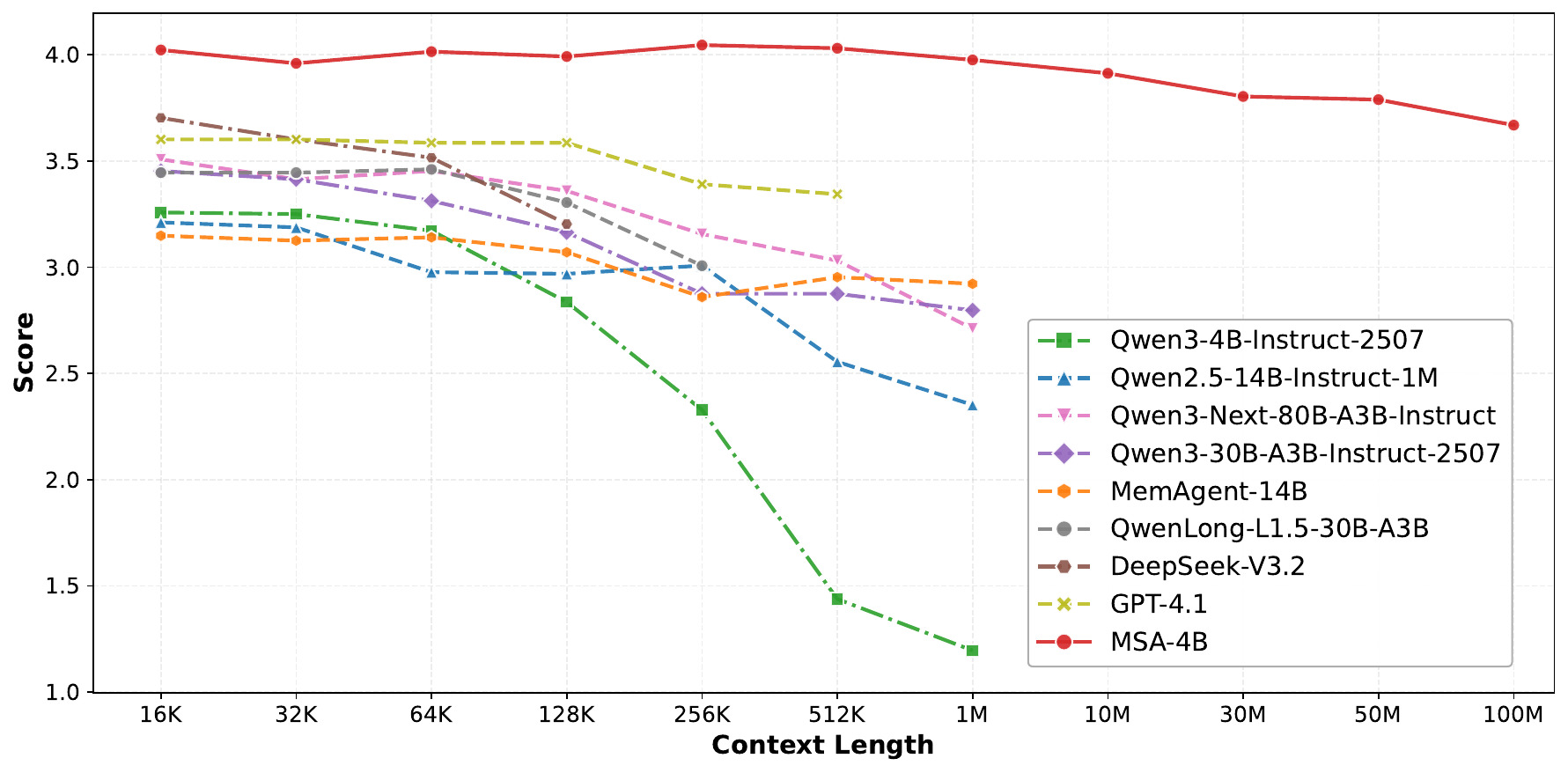}
    \caption{ MSA integrates top$k$ selection with sparse attention, achieving strong scalability while remaining differentiable. This design enables end-to-end training, yet allows the documents to be decoupled at inference time, thereby providing robust extrapolation capability. MSA demonstrates exceptional scalability on the MS MARCO dataset, sustaining consistent performance with less than 9\% degradation across an unprecedented memory context range from 16K to 100M tokens. Some curves terminate prematurely due to context length limitations.}
    \label{fig:scaling}
\end{figure}

An effective long-term memory system for LLMs should satisfy several core desiderata: seamless compatibility with mainstream model architectures, scalability to lifetime memory with low computational overhead and minimal degradation in model quality, end-to-end trainable mechanisms that enable high-precision retrieval and storage, straightforward memory management, and robustness against catastrophic forgetting.

As summarized in Table~\ref{tab:recentwork}, current paradigms for LLM memory fall into three principal categories, each addressing only a subset of the essential criteria for scalable, high-fidelity lifelong memory. \textbf{(I) Parameter-Based Memory} internalizes new knowledge by directly updating model parameters (e.g., LoRA~\cite{LoRA}, Continual Pre-training) or leveraging learnable architectures adapted via test-time training  (e.g., Titans~\cite{behrouz2024titans}). Although these methods offer strong architectural compatibility and deep semantic integration with high precision, they fundamentally lack capacity scalability: parameter updates are vulnerable to catastrophic forgetting, particularly under conflicting knowledge, and incur significant training overhead with complex memory management. \textbf{(II) External Storage-Based Memory}, typified by Retrieval-Augmented Generation (RAG) and MemAgent, retrieves relevant information from large external knowledge stores. This paradigm preserves base model capabilities, scales naturally to lifetime-sized memory banks, and avoids catastrophic forgetting. However, its reliance on discrete semantic representations (e.g., raw text or embeddings) prevents end-to-end differentiability. The resulting decoupled retrieval pipeline imposes an intrinsic performance ceiling, limiting these systems to medium precision and shallow semantic matching that aligns only weakly with the model’s internal reasoning space. \textbf{(III) Latent State-Based Memory} aims to construct memory directly from internal latent representations (e.g., hidden states or KV caches), offering high semantic fidelity by operating within the model’s native representation space. Yet this approach introduces a strict trade-off between capacity and efficiency. KV-centric methods (e.g., DSA \cite{liu2025deepseek}, MemGen \cite{zhang2025memgenweavinggenerativelatent}) maintain strong precision and architectural compatibility but incur prohibitive computational costs, preventing them from scaling to extreme 100M-token contexts. Conversely, linear-attention-based variants (e.g., RWKV \cite{peng2023rwkv}, DeltaNet \cite{yang2024parallelizing}) achieve efficient $\mathcal{O}(L)$ complexity by recurrently compressing history into fixed-size states. However, their bounded capacity inevitably causes catastrophic forgetting under extreme-length settings, severely degrading precision and reducing architectural alignment with mainstream LLMs.

\begin{table}[ht]
\centering
\small
\caption{Comparison of Long-Term Memory Methods for LLMs}
\setlength{\tabcolsep}{3pt}
\renewcommand{\arraystretch}{1.15}

\begin{tabularx}{\linewidth}{
    ccccccc
}
\toprule
Method &
\makecell[c]{Lifetime\\ Memory} &
Precision &
\makecell[c]{Compatible w/\\Mainstream LLMs} &
\makecell[c]{Computational\\Complexity} &
\makecell[c]{Memory\\Management} &
\makecell[c]{Catastrophic \\Forgetting}\\
\midrule
\multicolumn{7}{l}{\textit{\textbf{Parameter-Based Memory}}} \\
\makecell[c]{Model-Based \\(LoRA / CPT)} & \color{Red}{No} & \color{Green}{High} & \color{Green}{High} &  \makecell[c]{Training: \color{Red}{High}\\ Inference: \color{Green}{Low}} & \color{Red}{Hard} & \color{Red}{Yes} \\

\makecell[c]{Test-Time Training\\ (Titans)} & \color{Red}{No} & Medium & \color{Red}{Low} & Medium & Medium & \color{Red}{Yes} \\
\midrule
\multicolumn{7}{l}{\textit{\textbf{External Storage-Based Memory}}} \\
RAG & \color{Green}{Yes} & Medium & Medium & \color{Green}{$\mathcal{O}(L)$} & \color{Green}{Easy} & Low \\

MemAgent & \color{Green}{Yes} & Medium & Medium & Medium & \color{Green}{Easy} & Medium \\
\midrule
\multicolumn{7}{l}{\textit{\textbf{Latent State-Based Memory}}} \\
\makecell[c]{Sparse Attention\\(DSA)} & \color{Red}{No} & \color{Green}{High} & \color{Green}{High} & Medium & \color{Green}{Easy} & \color{Green}{No} \\

\makecell[c]{Linear Attention\\(DeltaNet/RWKV)} & \color{Red}{No} & \color{Red}{Low} & \color{Red}{Low} & \color{Green}{$\mathcal{O}(L)$} & \color{Green}{Easy} & \color{Red}{Yes} \\

MemGen & \color{Red}{No} & Medium & \color{Green}{High} & Medium & Medium & \color{Green}{No} \\
% \midrule
\noalign{\vskip 0.5ex}
\hdashline
\noalign{\vskip 0.5ex}
% \rowcolor{gray!15}
% \rowcolor{blue!5}
\textbf{MSA (Ours)} & \color{Green}{\textbf{Yes}} & \color{Green}{\textbf{High}} & \color{Green}{\textbf{High}} & \color{Green}{\textbf{$\mathcal{O}(L)$}} & \color{Green}{\textbf{Easy}} & \color{Green}{\textbf{No}} \\
\bottomrule
\end{tabularx}

\label{tab:recentwork}
\end{table}

Overall, existing approaches remain constrained by two fundamental limitations: \textbf{(I) limited scalability of high-fidelity memory.} Methods that deliver strong precision are bound by fixed context or state capacity, while methods that scale in capacity struggle to ensure reliable effectiveness. \textbf{(II) lack of end-to-end trainability.} No current paradigm offers a fully differentiable, jointly optimized memory pipeline that simultaneously preserves architectural compatibility, high precision, and robustness against catastrophic forgetting across all scales.

To address these challenges, we propose Memory-Sparse Attention (MSA), a novel, end-to-end trainable, and scalable sparse attention mechanism designed specifically for lifelong memory contexts.
As a latent state-based approach, MSA integrates top-$k$ selection with sparse attention, achieving strong scalability while remaining differentiable.
By leveraging KV cache sparsification, MSA achieves near-linear time complexity and supports inference over 100M tokens through optimized implementation.
Furthermore, we introduce a global and document-wise Rotary Positional Embedding (RoPE) mixed strategy to extend the context window.
This design allows MSA to be trained efficiently on 64k contexts while effectively extrapolating to 100M tokens, significantly reducing training overhead.
Experimental results demonstrate that MSA achieves state-of-the-art (SOTA) performance on long-text Question Answering tasks,
outperforming baseline models with identical backbones and surpassing advanced RAG systems on most benchmarks.
Additionally, MSA achieves SOTA results on the "Needle-In-A-Haystack" (NIAH) test, exhibiting superior robustness against context degradation.

As illustrated in Figure~\ref{fig:scaling}, MSA demonstrates unprecedented scalability, maintaining performance with less than 9\% degradation across context ranges spanning from 16K to 100 million tokens, which is a scale approaching the estimated capacity of human lifelong memory. In comparison, traditional long-context models (e.g., Qwen2.5-14B-1M~\cite{qwen2.5}, Qwen3-30B/80B-A3B~\cite{qwen3-coder}) and external memory systems (e.g., MemAgent-14B~\cite{yu2025memagent}) suffer from catastrophic degradation at this scale. Unlike SOTA RAG systems, MSA eliminates the need for complex retrieval pipelines and heuristic hyperparameters, such as top-k recall or relevance thresholds. This capability marks significant progress in bridging the gap between LLM memory and human cognitive scale, enabling practical applications previously deemed unattainable for neural models.

Our contributions are summarized as follows:
\begin{itemize}
    \item We propose MSA, an end-to-end trainable, scalable sparse attention architecture with a document-wise RoPE that extends intrinsic LLM memory while preserving representational alignment. It achieves near-linear inference cost and exhibits $<9\%$ degradation even when scaling from 16K to 100M tokens.
    \item We introduce KV cache compression to reduce memory footprint and latency while maintaining retrieval fidelity at scale. Paired with Memory Parallel, it enables high-throughput processing for 100M tokens under practical deployment constraints, such as a single $2\times \text{A}800$ GPU node.
    \item We present Memory Interleave, an adaptive mechanism that facilitates complex multi-hop reasoning. By iteratively synchronizing and integrating KV cache across scattered context segments, MSA preserves cross-document dependencies and enables robust long-range evidence integration.
    \item Comprehensive evaluations on long-context QA and Needle-In-A-Haystack benchmarks demonstrate that MSA significantly outperforms frontier LLMs, state-of-the-art RAG systems and leading memory agents.
\end{itemize}

\section{Related Work}

As outlined in the introduction, recent research on augmenting LLMs with memory capabilities generally falls into three paradigms.

\textbf{Parameter-based memory.} This paradigm seeks to internalize external information directly into the model's parameters. A foundational approach involves direct fine-tuning on domain-specific data using techniques such as Continuous Pre-training (CPT) or LoRA. This strategy is widely adopted to embed procedural knowledge and reasoning patterns~\cite{chen2023fireactlanguageagentfinetuning, yin2024agentlumosunifiedmodular, zhang2024agentohanadesignunifieddata, fu2025agentrefineenhancingagentgeneralization}. To mitigate catastrophic forgetting and decouple memory from reasoning, recent research has shifted towards specialized architectural components. MLP-Memory~\cite{wei2025mlpmemoryretrieverpretrainedmemory}, for instance, substitutes explicit retrieval with a parametric retriever, training an MLP to act as a differentiable memory store. Scaling this modular concept further, FLEXOLMO~\cite{shi2025flexolmoopenlanguagemodels} introduces a mixture-of-experts framework that updates specific modules for targeted knowledge integration, while Engram~\cite{cheng2026conditional} augments the model with massive sparse memory structures via N-gram embeddings to bypass the capacity bottlenecks of dense layers. Pushing the paradigm towards "dynamic neural memory," recent innovations such as Titans~\cite{behrouz2024titans} and Nested Learning~\cite{behrouz2025nested} propose maintaining memory modules whose weights are updated during inference (test-time training), treating context processing as a nested optimization loop. This direction is theoretically grounded in frameworks like MIRAS~\cite{behrouz2025s}, which unifies such recurrent and associative memory architectures under a common abstraction.

\textbf{External storage-based memory.} This paradigm augments models with a large-scale external database, from which relevant memories are extracted via semantic retrieval on demand. The foundational framework in this category is Retrieval-Augmented Generation (RAG)~\cite{NEURIPS2020_RAG}, which retrieves textual chunks based on vector similarity between the query and the external corpus. To address the precision limitations of initial dense retrieval, which can introduce irrelevant or "noisy" context, state-of-the-art RAG systems frequently incorporate a reranking stage to refine the candidate list, ensuring that only the most pertinent information occupies the model's limited context window. Recent innovations have sought to optimize the format of retrieved memory. Memory³~\cite{memory3}, for instance, pre-encodes external knowledge into structured KV pairs for direct injection into the model's attention layers. Crucially, however, the retrieval process in Memory³ remains grounded in model-agnostic semantic embeddings rather than the model's internal state, maintaining an optimization gap between the retrieval metric and the generation objective. To bridge this gap, MemAgent~\cite{yu2025memagent} formulates memory management as a sequential decision-making process. By employing Reinforcement Learning, it trains the model to actively read, write, and overwrite memory segments, thereby aligning the information retention policy directly with the downstream reasoning performance rather than relying solely on static similarity metrics. Addressing the structure of memory, MemGAS~\cite{xu2025singlemultigranularitylongtermmemory} improves upon the flat indexing of standard RAG by introducing a hierarchical management mechanism. This allows for multi-granularity retrieval, enabling the system to adaptively fetch information ranging from coarse-grained summaries to fine-grained details depending on the specific query requirements.

\textbf{Latent state-based memory.} Distinct from model-agnostic semantic retrieval-based memory, the latent memory paradigm constructs and manages memory directly using the model's internal latent states. As noted previously, Memory³ attempts to leverage this by encoding information into KV pairs; however, constrained by the prohibitively large size of active KV caches, it offloads these representations to an external database. Consequently, it still relies on model-agnostic semantic embeddings as retrieval keys to concatenate retrieved pairs with the context, rather than maintaining a persistent internal state. In contrast, more intrinsic approaches aim to manage the model's working memory directly. ParallelComp~\cite{xiong2025parallelcompparallellongcontextcompressor} addresses the capacity limit by implementing sophisticated KV cache eviction policies to dynamically compress context during inference. Similarly, MemGen~\cite{zhang2025memgenweavinggenerativelatent} exploits the model's autoregressive capabilities to iteratively synthesize and compress historical information into compact memory representations, thereby retaining essential information within the model's latent space.

Another distinct class of latent memory is Linear Attention mechanisms. In contrast to standard attention, which requires explicit access to previous KV, linear attention naturally compresses information from the preceding sequence into compact hidden states during the recurrence. Architectures such as RWKV~\cite{peng2023rwkv} formulate attention as a linear recurrence (WKV), where historical context is aggregated into a time-decaying hidden state. Similarly, DeltaNet~\cite{schlag2021linear,yang2024parallelizing} updates its memory state using a delta rule, iteratively refining value representations based on new inputs. While compressing the entire history into fixed-size latent states yields substantial computational and storage efficiency, it inherently involves lossy compression. Consequently, when constrained by a finite state size, these methods inevitably suffer from severe performance degradation and information loss as the memory context extends to extreme-long scales.
\section{Memory Sparse Attention}
\subsection{Overall Design}
We introduce \textbf{MSA} (Memory Sparse Attention), a unified, end-to-end trainable latent memory framework designed for massive memory Question-Answering. The core principle of MSA is to seamlessly integrate the processes of Memory sparse retrieval and answer generation into a single, jointly-optimized architecture, moving beyond the limitations of conventional decoupled "retrieve-then-read" pipelines while preserving the ability to handle long-context memory. 

\subsection{Architecture}

\subsubsection{Sparse Attention Mechanism}

\begin{figure*}[!h]
\vskip 0.2in
\begin{center}
\centerline{\includegraphics[width=1\columnwidth]{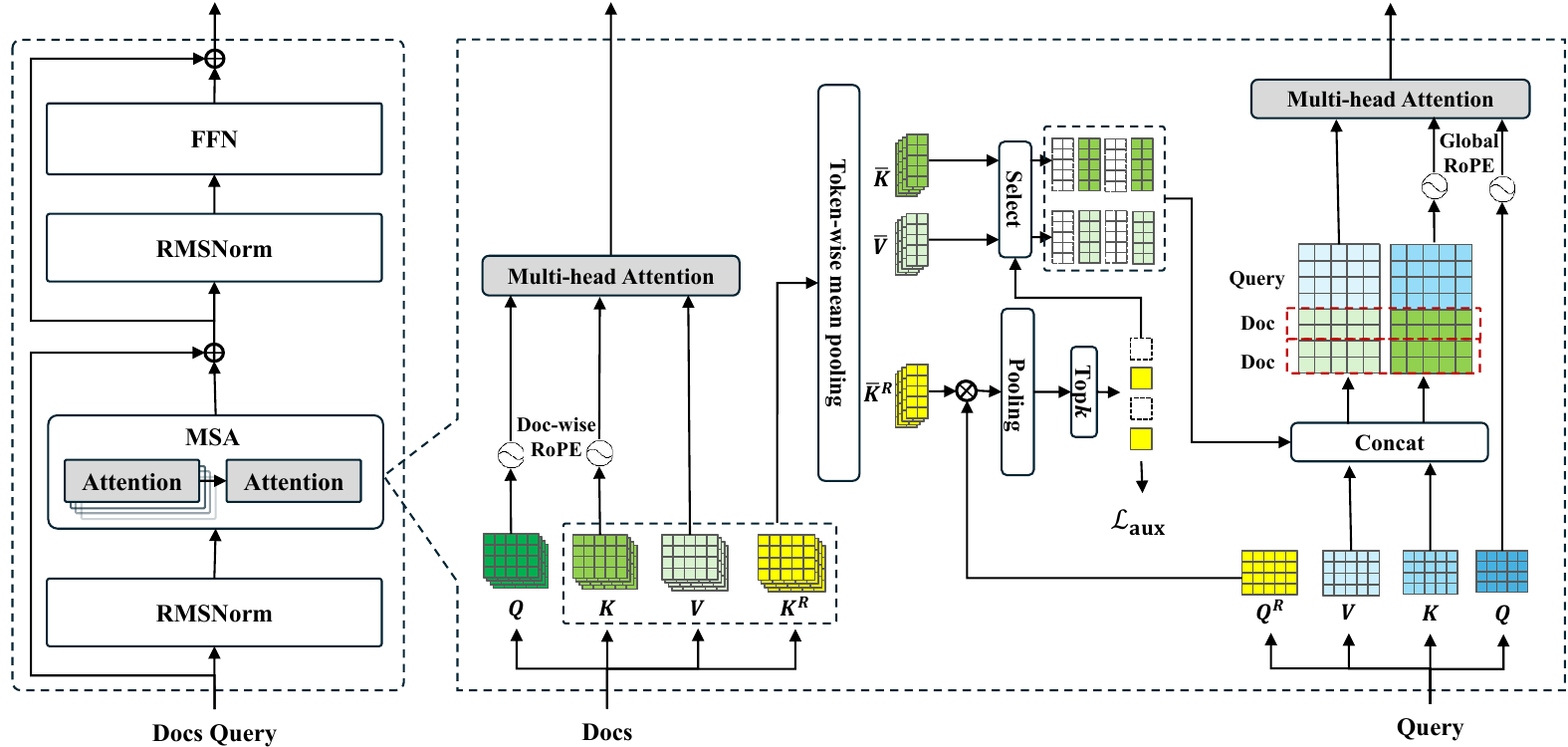}}
\caption{Memory Sparse Attention layer}
\label{fig:EverMemModel} 
\end{center}
\vskip -0.2in
\end{figure*}

As shown in Figure~\ref{fig:EverMemModel}, to efficiently process massive memory at the latent state level, MSA replaces the standard dense self-attention with a document-based retrieval sparse attention mechanism. Formally, let the memory bank consist of a set of documents $\mathcal{D} = \{d_1, d_2, \dots, d_N\}$. For each document $d_i$, let $H_i$ denote its hidden state representation. For a specific attention head $h$, we generate the standard Key $K_{i,h}$ and Value $V_{i,h}$ matrices via the backbone model's projection weights $W^h_K$ and $W^h_V$. In parallel, we introduce a \textit{Router K Projector}, parameterized by $W^h_{K^R}$, to generate a specialized routing key matrix $K^R_{i,h}$:
\begin{equation}
K_{i,h} = H_i W^h_K, \quad V_{i,h} = H_i W^h_V, \quad K^R_{i,h} = H_i W^h_{K^R}.
\end{equation}
To significantly reduce the memory footprint and retrieval complexity, we segment each document into multiple fixed‑length chunks and perform chunk‑wise mean pooling, denoted as $\phi(\cdot)$, to compress these states into latent representations. This yields the compressed matrices $\bar{K}_{i,h} = \phi(K_{i,h})$, $\bar{V}_{i,h} = \phi(V_{i,h})$, and $\bar{K}^R_{i,h} = \phi(K^R_{i,h})$.

During inference, given a user query with hidden state $H_q$, for a specific attention head, we similarly compute its standard states $Q_{q,h}, K_{q,h}, V_{q,h}$ via the backbone's $W^h_Q, W^h_K, W^h_V$ projections. Simultaneously, a  \textit{Router Q Projector} $W^h_{Q^R}$ generates a specific routing query $Q^R_{q,h} = H_q W^h_{Q^R}$.
The relevance score $S_{ij}$ for the $j$-th chunk of the $i$-th document is computed as the cosine similarity between the query's routing vector $Q^R_{q,h}$ and the memory's compressed routing keys $\bar{K}^R_{ij,h}$,
and is first aggregated across attention heads using mean pooling.
To identify the most relevant memory segments, a maximum pooling is then applied over the query‑token–level relevance scores, i.e.,
\begin{equation}
S_{ij} = \max_{\text{token } t} (\underset{\text{head }h}{\text{mean}}(\text{cos}((Q^R_{q,h})_{t}, \bar{K}^R_{ij,h}))),
\end{equation}
where $\text{cos}(\cdot)$ denotes cosine similarity. The document-level relevance score is defined as the maximum score among its constituent chunks, $s_i = \max_j S_{ij}$. Based on these scores, we select the indices of the Top-$k$ documents, denoted as $\mathcal{I} = \text{Top-}k(\{s_i\}_{i=1}^N)$. Finally, the generation is performed by concatenating the compressed Key and Value matrices of the selected documents before the query's local cache. The model then performs autoregressive generation where the query $Q_q$ from active tokens attends to this aggregated, sparsity-aware context:
\begin{equation}
K_{\text{ctx}} = [\{\bar{K}_i\}_{i \in \mathcal{I}}; K_q], \quad V_{\text{ctx}} = [\{\bar{V}_i\}_{i \in \mathcal{I}}; V_q],
\end{equation}
\begin{equation}
\text{Output} = \text{Attention}(Q_q, K_{\text{ctx}}, V_{\text{ctx}}).
\end{equation}

% \paragraph{Layer-wise Routing Allocation.}
We implement the MSA routing strategy selectively, applying it exclusively to the latter half of the model's layers. Empirical analysis reveals that the hidden states in the initial layers fail to capture the high-level semantic abstractions necessary for effective retrieval, rendering the routing mechanism inefficient at these depths. Consequently, in the lower layers (without MSA routing), while we retain Independent Document Processing to update document states and ensure hierarchical representation alignment, we bypass the sparse retrieval and memory integration steps. In these layers, the generation process relies solely on the local context, without attending to the compressed memory KV pairs.

\label{subsec:architecture}

\subsubsection{Parallel and Global RoPE}

To ensure robust generalization across varying memory scales, MSA employs independent RoPE for each document. A critical challenge in scaling memory is the discrepancy between training and inference contexts: models are typically trained with a limited number of documents due to compute constraints, i.e., train-on-short, but must operate on massive document banks during inference, i.e., infer-on-long.

Standard global positional encodings would assign monotonically increasing position IDs across the concatenated sequence~\cite{su2024roformer}. This causes the position indices to shift drastically as the number of documents grows, leading to severe performance degradation when the inference context length exceeds the training horizon. By assigning independent position IDs (starting from 0) to each document, MSA decouples the positional semantics from the total number of documents in memory.  Consequently, the model can effectively extrapolate, maintaining high retrieval and reasoning accuracy on massive memory contexts even after being trained only on smaller subsets.

Complementing this parallel strategy, we employ Global RoPE for the active context, which includes the user query and the subsequent autoregressive generation. The position IDs for these tokens are offset by the number of retrieved documents. Specifically, the position indices for the query initiate from $k$ (corresponding to the Top-$k$ retrieved compressed KVs). This strategic offset ensures that the model perceives the active context as a logical continuation of the retrieved background information, thereby preserving the causal dependency essential for coherent generation.

\subsection{Training}
\label{subsec:training}

\subsubsection{Continuous Pre-training}

To endow the model with robust retrieval capabilities, we perform continuous pre-training on a deduplicated corpus comprising 158.95 billion tokens. The overarching objective of this stage is to train the model to perform Generative Retrieval, where the model autoregressively generates the unique document IDs of relevant documents.

To explicitly guide the internal sparse attention mechanism beyond the supervision provided by the standard generation loss $\mathcal{L}_{\text{LLM}}$, we introduce an auxiliary loss, $\mathcal{L}_{\text{aux}}$, designed to supervise the Layer-wise Routing process. Within each MSA layer, the Router Projector is responsible for selecting the Top-$k$ most relevant documents to participate in attention. We apply $\mathcal{L}_{\text{aux}}$ to these intermediate routing decisions to ensure that the model attends to the correct evidence. 
Inspired by \cite{khosla2020supervised}, for a given query $q$, let $\mathcal{D}$ denote the associated document set, and let $\mathcal{P}\subseteq \mathcal{D}$ be the set of positive documents.
The set of negative documents is $\mathcal{N}=\mathcal{D}\setminus \mathcal{P}$, whose cardinality is $|\mathcal{N}|=|\mathcal{D}|-|\mathcal{P}|$.
Let $s_i^{+}$ denote the relevance score of the $i$-th positive query--document pair and $s_{i,j}^{-}$ the relevance score of the $j$-th negative paired with the $i$-th positive. 
The auxiliary loss is then defined as:
\begin{equation}
\label{eq:aux-loss-indexed}
\mathcal{L}_{\text{aux}}
= - \frac{1}{|\mathcal{P}|} \sum_{i=1}^{|\mathcal{P}|}
\log \frac{
\exp\!\big( s_i^{+} / \tau \big)
}{
\exp\!\big( s_i^{+} / \tau \big)
+ \sum_{j=1}^{|\mathcal{N}|} \exp\!\big( s_{i,j}^{-} / \tau \big)
},
\end{equation}
where $\tau$ is the temperature parameter.
This supervised contrastive objective explicitly enforces separation between relevant and irrelevant document chunks in the latent routing space.

To ensure stability, we adopt a two-phase optimization schedule. In the initial warm-up phase, we focus on aligning these internal Router Projectors. We set the total loss to $\mathcal{L} = 0.1 \mathcal{L}_{\text{LLM}} + \mathcal{L}_{\text{aux}}$ with a learning rate of $1\text{e-}4$. This encourages the router heads to quickly learn effective selection policies. Upon completion of the warm-up, we transition to the main pre-training phase, where the learning rate is annealed to $6\text{e-}6$ and the loss weights are adjusted to $\mathcal{L} = \mathcal{L}_{\text{LLM}} + 0.1 \mathcal{L}_{\text{aux}}$. This configuration prioritizes the ultimate Generative Retrieval task while maintaining the discriminative power of the internal layer-wise routing established during warm-up.

\subsubsection{Post-Training}

Following continuous pre-training, we implement a two-stage curriculum learning strategy for SFT on Question Answering tasks. 

In the first stage, we conduct SFT on a large-scale dataset with a context length of 8k tokens. The primary objective of this phase is to establish the model's fundamental instruction-following and reasoning capabilities within a standard context window. 

In the second stage, we focus on enhancing data quality and length extrapolation. We apply a rigorous data cleaning process to filter out erroneous and low-quality samples from the training set. Concurrently, we extend the memory context length from 8k to 64k tokens. This curriculum transition enables the model to adapt to longer dependencies and significantly improves its robustness when extrapolating to massive memory banks during inference.

\subsection{Inference}
\label{subsec:inference}

\subsubsection{Three-Stage Inference Process}
\begin{figure*}[!h]
\vskip 0.2in
\begin{center}
\centerline{\includegraphics[width=1\columnwidth]{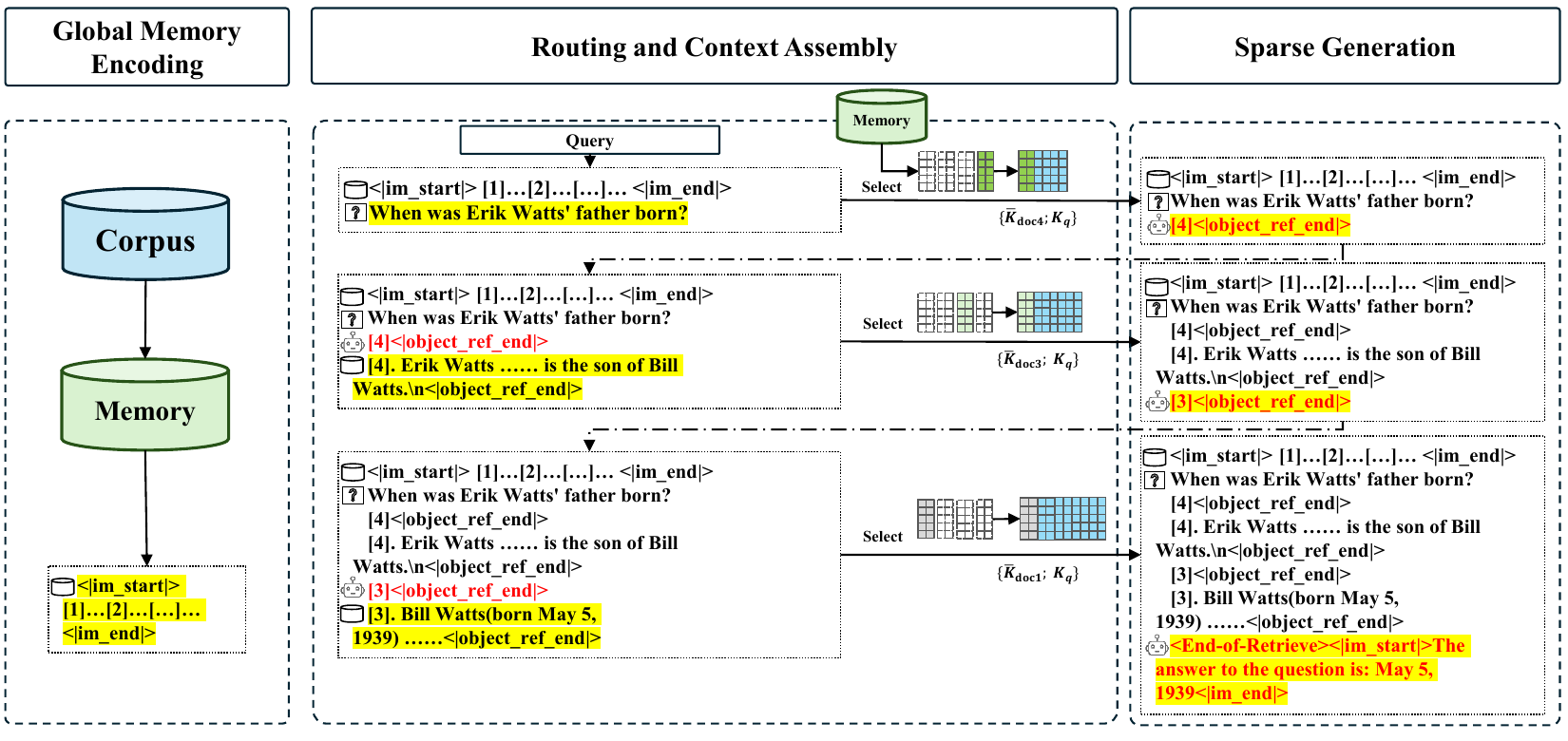}}
\caption{Three-Stage Inference Process with Memory Interleave}
\label{fig:ModelInference} 
\end{center}
\vskip -0.2in
\end{figure*}

% To process this massive volume of documents with high efficiency, we design a optimized three-stage inference pipeline that decouples the heavy memory encoding from the lightweight query processing.
The inference pipeline is designed to handle the large-scale memory bank efficiently through three distinct stages, as shown in Figure~\ref{fig:ModelInference}:

\textbf{Stage 1: Global Memory Encoding (Offline).}
This stage is a one-time, offline pre-computation over the entire document corpus. For every document, the model performs a forward pass to generate the standard  $K$ and $V$ matrices. Simultaneously, the specialized \textit{Router K Projector} generates the routing key matrix $K^R$. To minimize storage and retrieval latency, all three matrices ($K, V, K^R$) are partitioned into chunks and compressed via mean pooling. The resulting compact representations ($\bar{K}, \bar{V}, \bar{K}^R$) are then cached in the memory bank. This stage converts the raw text corpus into a structured, retrievable latent store.

\textbf{Stage 2: Routing and Context Assembly (Online).}
This stage is initiated upon receiving a user question. First, the model computes the question's hidden states and projects them via the \textit{Router Q Projector} to obtain the routing query $Q^R_q$. This query is then matched against the cached global routing keys $\bar{K}^R$ to calculate relevance scores and identify the Top-$k$ documents. Crucially, strictly for the attention mechanism, only the compact Key and Value matrices ($\bar{K}, \bar{V}$) of these selected documents are loaded. These are then concatenated with the question's local $K_q$ and $V_q$ to form the final sparse context.

\textbf{Stage 3: Sparse Generation (Online).}
In the final stage, the model operates autoregressively on the assembled sparse context. The standard attention mechanism computes the interaction between the active token's Query $Q_q$ and the concatenated KV pairs $[\{\bar{K}_{topk}\}; K_q]$, generating the final answer token by token.

\subsubsection{Memory Parallel}

We have designed a specialized inference engine to enable extreme‑length memory inference on a standard single node.
% equipped with 8 NVIDIA A100 GPUs.
Under this engine, MSA supports inference over a massive memory context of up to \textbf{100 million tokens} with only 2 NVIDIA A800 GPUs.
Operating on such a massive context scale presents significant challenges in terms of memory capacity and computational efficiency. To address these, we implement a tailored optimization pipeline, Memory Parallel, covering the entire lifecycle from encoding to retrieval.

\textbf{Tiered Memory Storage Strategy.}
For the runtime storage, a theoretical estimation indicates that the compressed KV and $\bar{K}^R$ cache for 100M tokens, assuming a pooling kernel size $P=64$, 8 heads, and head dimension 128 across 18 layers in BF16, would require approximately 169GB of memory. This figure strictly exceeds the aggregate 160GB capacity of a standard 2$\times$A800 node, rendering a monolithic storage approach physically impossible, even before accounting for the memory required by model parameters and dynamic activation overheads. We observe that retrieval only requires the routing keys $\bar{K}^R$, while the content $\bar{K}$ and $\bar{V}$ are needed only after selection. Thus, we design a tiered storage system:

\begin{itemize}
    \item \textbf{GPU-Resident Routing Keys:} To ensure low-latency retrieval, we distribute the Routing Keys ($\bar{K}^R$) across the VRAM of multiple GPUs. Even with optimizations, $\bar{K}^R$ alone can occupy $\sim$56GB for a 100M context, necessitating distributed storage.
    \item \textbf{CPU-Offloaded Content KVs:} The bulk of the memory bank, the Content KVs ($\bar{K}, \bar{V}$), is stored in the host DRAM (CPU memory). Upon identifying the Top-$k$ relevant chunks via GPU scoring, only the corresponding Content KVs are asynchronously fetched from the host to the GPU for the subsequent attention computation.
\end{itemize}
This separation decouples the capacity requirement from VRAM limits, enabling 100M-token scale on standard hardware.

\textbf{Memory-Parallel Retrieval and Distributed Scoring.}
To address computational efficiency, we adopt a Memory Parallel strategy for retrieval. Given the relatively compact size of the 4B backbone, we replicate the full model weights on each GPU to avoid communication overhead during decoding. During the retrieval step, the query hidden states are broadcast to all GPUs. Each GPU independently calculates similarity scores against its local shard of Routing Keys. These scores are then reduced globally to identify the Top-$k$ indices. Additionally, we implement a tiling mechanism for the scoring matrix multiplication to manage peak memory usage and prevent OOM errors during these large-scale operations.

\subsection{Memory Interleave}

To address complex queries requiring multi-hop reasoning, MSA incorporates an adaptive Memory Interleave Mechanism that essentially performs the routing and context assembly (Stage 2) and Sparse Generation (Stage 3) in an iterative manner.
Unlike single-shot retrieval, the inference process alternates between Generative Retrieval and Context Expansion, in which the retrieved documents will be treated as a part of the query for the next iteration, as shown in Figure~\ref{fig:ModelInference}.
After loading the KV-cache for the document corpus, the model first autoregressively generates a sequence of document IDs ending with a special delimiter based on the given query.
Note that the number of documents generated in each round is not fixed, but is adaptively determined by the model.
Once the document IDs are generated, the system obtains the corresponding original texts and appends them to the original query, which is then leveraged in the next iteration.
This cycle, which generates evidence identifiers, retrieves global context, and updates the state, repeats adaptively until the model determines that the accumulated documents are sufficient, at which point it transitions from generating document IDs to autoregressively generating the final answer.

Notably, under the inference design described above, each retrieval chain in the multi-hop datasets is divided into multiple training samples during model training.
Each sample contains a single retrieval step, either based on the single query or on the existing document context, and samples are randomly selected for training.

\section{Experiment}
\subsection{Experimental Setup}
\textbf{Overview.}
To comprehensively evaluate the efficacy of MSA, we conduct experiments on both Question Answering (QA) task and long-context "Needle In A Haystack" (NIAH) task. For the QA task, we assess performance on nine standard benchmarks. To ensure a rigorous comparative analysis, we benchmark MSA against two types of Retrieval-Augmented Generation (RAG) systems: a controlled baseline built upon the identical Qwen3-4B-Instruct-2507 backbone to isolate the architectural contributions of MSA, and a "best-of-breed" baseline composed of State-of-the-Art (SOTA) modules for each component to test against peak performance. In the NIAH domain, we utilize the RULER dataset~\cite{hsieh2024ruler} to evaluate long-context fidelity. Here, we compare our approach against both external storage-based memory systems and latent state-based memory architectures.

\textbf{Datasets and Metrics.}
We evaluate MSA on nine diverse benchmarks covering single-hop, multi-hop, and long-context scenarios: MS MARCO v1, Natural Questions, DuReader, TriviaQA (10M), NarrativeQA, PopQA, 2WikiMultiHopQA, HotpotQA, and MuSiQue, with memory banks ranging from 277K to 10M tokens. For the standard RAG systems, we report performance metrics (LLM judge, whose prompt is shown in Appendix~\ref{sec:appendixA}) at fixed retrieval depths of $k=\{1, 5, 10\}$, denoted as \textbf{R@1, R@5, and R@10}, respectively. Notably, for the RAG systems that perform retrieval and reranking, we first retrieve 100 candidate documents and then select the top‑$\{1, 5, 10\}$ items based on the reranked list. 
In contrast, for our MSA models, we utilize an \textbf{@adaptive} metric. This indicates that instead of relying on a pre-defined, fixed number of retrieved documents, the model autonomously determines the number of documents required to answer each specific query.
For NIAH evaluation, we employ the RULER benchmark, which consists of eight diverse sub-tasks covering both standard single-needle retrieval (SA1-3) and complex multi-needle scenarios involving multiple keys, values, and queries (MK1-3, MV, MQ). We report the average accuracy across these tasks to comprehensively assess the model's stability and extrapolation capabilities from 32K up to 1M tokens.

\textbf{Implementation Details.}
Our MSA model is built upon the Qwen3-4B-Instruct-2507 architecture. We initialize the backbone parameters using the official pre-trained weights to leverage its established capabilities, while the newly introduced router projectors are randomly initialized. The model undergoes continuous pre-training on the 158.95B token corpus. To ensure stability, we employ the two-stage pre-training schedule described in Sec.~\ref{subsec:training}, transitioning from a retrieval-focused warmup ($\mathcal{L} = 0.1 \mathcal{L}_{\text{LLM}} + \mathcal{L}_{\text{aux}}$) to the main pre-training phase ($\mathcal{L} = \mathcal{L}_{\text{LLM}} + 0.1 \mathcal{L}_{\text{aux}}$). Regarding the specific MSA hyperparameters, we set the compression chunk size to 64 tokens and configure the router to select the Top-16 relevant documents for attention. We evaluate two model variants to analyze the impact of our curriculum learning strategy: MSA-S1, which is fine-tuned solely through the first stage of post-training with a standard 8k context; and MSA-S2, which undergoes the complete two-stage curriculum learning pipeline, extending the memory context to 64k.

\textbf{Baselines.}
We evaluate MSA on QA task and NIAH task with task-specific baselines for each. For QA task, we compare against two categories of RAG baselines. (I) same-backbone RAG: MSA is initialized with Qwen3-4B-Instruct-2507~\cite{qwen3-coder}. To validate the effectiveness of MSA, we evaluate RAG systems built on the same backbone, including standard RAG (Qwen3-4B-Embedding~\cite{zhang2025qwen3embeddingadvancingtext} + Qwen3-4B-Instruct-2507), RAG with reranking (adding Qwen3-4B-Rerank), and HippoRAG2~\cite{gutierrez2025rag}, a knowledge graph-augmented RAG framework. (II) best-of-breed RAG: We further compare against state-of-the-art configurations employing KaLMv2-Embedding-Gemma3-12B-2511~\cite{zhao2025kalm} as the retriever, paired with frontier-scale generators including Qwen3-235B-Instruct-2507 and Llama-3.3-70B-Instruct~\cite{llama3}, with optional reranker Qwen3-8B-Rerank. Unless otherwise noted, all standard RAG pipelines are implemented using UltraRAG v2.0~\cite{chen2025ultrarag}. For NIAH task, we compare against external storage-based memory method MemoryAgent-14B~\cite{Yu2025MemAgentRL} and mixed linear attention models, including Qwen3-Next-80B-A3B, Qwen3-30B-A3B, Qwen2.5-14B-1M. Additionally, our backbone model Qwen3-4B-Instruct-2507 is also included.

\subsection{Main Results}
\subsubsection{QA task}

\begin{table*}[!h]
\centering
\caption{
Comparison of MSA with same-backbone RAG baselines (Qwen3-4B) on LLM judge results (scale 0-5). Higher scores indicate better performance. "RR" denotes RAG systems that perform both retrieval and reranking.
}
\label{tab:qa_results_4b}

\resizebox{1\linewidth}{!}{
\begin{tabular}{lr ccc ccc ccc c} 
\toprule
 & \textbf{Tokens} & \multicolumn{3}{c}{\textbf{Qwen3-4B}} & \multicolumn{3}{c}{\textbf{Qwen3-4B (RR)}} & \multicolumn{3}{c}{\textbf{Hipporag2}} & \textbf{MSA (Ours)} \\
\cmidrule(lr){3-5} \cmidrule(lr){6-8} \cmidrule(lr){9-11} \cmidrule(lr){12-12}
\textbf{Dataset} & & \textbf{R@1} & \textbf{R@5} & \textbf{R@10} & \textbf{R@1} & \textbf{R@5} & \textbf{R@10} & \textbf{R@1} & \textbf{R@5} & \textbf{R@10} & \textbf{@adaptive} \\
\midrule
MS MARCO v1~\cite{bajaj2016ms}      & 7.34M & 2.893 & 3.011 & 3.005 & 2.934 & \underline{3.032} & 3.017 & 2.676 & 3.005 & 3.019 & \textbf{4.141} \\
Natural Questions~\cite{47761}& 1.47M & 3.452 & 3.374 & 3.297 & \underline{3.494} & 3.408 & 3.385 & 3.338 & 3.389 & 3.374 & \textbf{3.545} \\
DuReader~\cite{he2018dureader}       & 277K  & 3.726 & 3.579 & 3.594 & \underline{3.848} & 3.618 & 3.607 & 2.941 & 3.485 & 3.415 & \textbf{4.155} \\
TriviaQA (10M)~\cite{Triviaqa}   & 10M   & 4.133 & 4.414 & 4.273 & 4.313 & 4.375 & 4.391 & 4.188 & \underline{4.430} & 4.367 & \textbf{4.621} \\
NarrativeQA~\cite{kovcisky2018narrativeqa}   & 538K  & 1.611 & 2.567 & 2.860 & \textbf{3.638} & 3.492 & \underline{3.536} & 1.959 & 2.628 & 2.655 & 3.395 \\
PopQA~\cite{mallen2023popqatrustlanguagemodelsinvestigating}         & 1.18M & 2.959 & 3.273 & 3.299 & \underline{3.315} & 3.264 & 3.266 & 3.111 & 3.249 & 3.249 & \textbf{3.433} \\
2WikiMultiHopQA~\cite{ho2020constructing2Wiki}  & 722K  & 1.065 & 3.055 & 3.136 & 1.187 & 3.057 & 3.159 & 1.045 & 3.180 & \underline{3.330} & \textbf{4.280} \\
HotpotQA~\cite{Hotpotqa}         & 1.35M & 2.252 & 3.582 & 3.787 & 2.642 & 3.990 & \underline{4.022} & 3.230 & 3.770 & 3.970 & \textbf{4.061} \\
MuSiQue~\cite{trivedi2022musique}        & 1.41M & 0.936 & 1.752 & 1.928 & 1.144 & 1.960 & 1.965 & 1.020 & 1.907 & \underline{2.095} & \textbf{2.211} \\
\midrule
\textbf{Average}  &       & 2.559 & 3.179 & 3.242 & 2.946 & 3.355 & \underline{3.372} & 2.612 & 3.227 & 3.275 & \textbf{3.760} \\
\bottomrule
\end{tabular}
}
\end{table*}

\begin{table*}[!h]
\centering
\caption{
Comparison of MSA with SOTA RAG systems using large-scale backbones on LLM judge results (scale 0-5). Higher scores indicate better performance. "RR" denotes RAG systems that perform both retrieval and reranking.
}
\label{tab:qa_results_sota}
\resizebox{1\linewidth}{!}{
\begin{tabular}{l C{0.95cm}C{0.95cm}C{0.95cm} C{0.95cm}C{0.95cm}C{0.95cm}
                C{0.95cm}C{0.95cm}C{0.95cm} C{0.95cm}C{0.95cm}C{0.95cm} c}
\toprule
 & \multicolumn{3}{c}{\textbf{KaLMv2 + Qwen3-235B}} 
 & \multicolumn{3}{c}{\textbf{KaLMv2 + Qwen3-235B (RR)}} 
 & \multicolumn{3}{c}{\textbf{KaLMv2 + Llama3.3}} 
 & \multicolumn{3}{c}{\textbf{KaLMv2 + Llama3.3 (RR)}} 
 & \textbf{MSA (Ours)} \\
\cmidrule(lr){2-4} \cmidrule(lr){5-7} \cmidrule(lr){8-10} \cmidrule(lr){11-13} \cmidrule(lr){14-14}
\textbf{Dataset} 
& \textbf{R@1} & \textbf{R@5} & \textbf{R@10} 
& \textbf{R@1} & \textbf{R@5} & \textbf{R@10} 
& \textbf{R@1} & \textbf{R@5} & \textbf{R@10} 
& \textbf{R@1} & \textbf{R@5} & \textbf{R@10} 
& \textbf{@adaptive} \\
\midrule
MS MARCO v1      & 2.846 & \underline{3.028} & 3.027 & 2.886 & 3.020 & 2.995 & 2.649 & 2.904 & 2.919 & 2.881 & 2.955 & 2.952 & \textbf{4.141} \\
Natural Questions & \underline{3.711} & 3.670 & 3.694 & 3.621 & 3.610 & 3.645 & 3.675 & 3.674 & 3.662 & \textbf{3.756} & 3.665 & 3.647 & 3.545 \\
DuReader          & \underline{4.044} & 3.991 & 3.978 & 3.973 & 3.932 & 3.891 & 4.051 & 3.846 & 3.742 & 3.967 & 3.776 & 3.780 & \textbf{4.155} \\
TriviaQA (10M)    & 4.367 & 4.656 & 4.578 & 4.492 & 4.320 & 4.555 & 4.273 & \textbf{4.740} & \underline{4.719} & 4.547 & 4.703 & 4.695 & 4.621 \\
NarrativeQA       & 1.413 & 2.130 & 2.427 & 3.212 & \textbf{3.427} & 3.375 & 1.290 & 2.123 & 2.382 & 3.150 & 3.263 & 3.317 & \underline{3.395} \\
PopQA             & 2.810 & 3.347 & \underline{3.396} & 3.268 & 3.380 & 3.376 & 2.787 & 3.298 & 3.305 & 3.337 & 3.384 & 3.362 & \textbf{3.433} \\
2WikiMultiHopQA   & 2.646 & 3.579 & 3.582 & 1.855 & 3.381 & \underline{3.583} & 1.339 & 3.263 & 3.445 & 1.651 & 3.332 & 3.541 & \textbf{4.280} \\
HotpotQA          & 3.497 & 4.090 & \textbf{4.225} & 3.341 & 4.141 & 4.194 & 3.070 & 3.896 & 4.127 & 3.428 & 4.145 & \underline{4.203} & 4.061 \\
MuSiQue           & 1.988 & 2.462 & \textbf{2.647} & 1.801 & 2.522 & 2.605 & 1.704 & 2.317 & 2.258 & 1.895 & 2.462 & \underline{2.614} & 2.211 \\
\midrule
\textbf{Average}  & 3.036 & 3.439 & 3.506 & 3.161 & 3.526 & \underline{3.580} & 2.760 & 3.340 & 3.396 & 3.179 & 3.521 & 3.568 & \textbf{3.760} \\
\bottomrule
\end{tabular}
}
\end{table*}

On the comprehensive suite of nine question answering benchmarks, MSA demonstrates consistent superiority over retrieval-augmented generation baselines constructed with the identical Qwen3-4B-Instruct backbone. Specifically, MSA achieves state-of-the-art performance on all datasets except NarrativeQA when compared against same-backbone RAG systems, as shown in Table~\ref{tab:qa_results_4b}. MSA achieves substantial performance gains, yielding average improvements of 16.0\%, 11.5\%, and 14.8\% over standard RAG, RAG with reranking, and HippoRAG2 (comparing against the best results among all recall settings), respectively. 

When benchmarked against best-of-breed RAG systems that integrate state-of-the-art components—including KaLMv2-Embedding paired with frontier-scale generators such as Qwen3-235B and Llama-3.3-70B—MSA secures top performance on four of the nine datasets while maintaining a competitive average score of 3.760. This represents relative improvements of 7.2\%, 5.0\%, 10.7\%, and 5.4\% over the strongest configurations of KaLMv2+Qwen3-235B, KaLMv2+Qwen3-235B (with reranking), KaLMv2+Llama-3.3, and KaLMv2+Llama-3.3 (with reranking), respectively. On the five datasets where MSA does not achieve absolute SOTA—Natural Questions, TriviaQA, NarrativeQA, HotpotQA, and MuSiQue—the performance gaps relative to the strongest baselines are 5.6\%, 2.5\%, 0.9\%, 3.9\%, and 16.5\%, respectively. Notably, for the multi-hop reasoning benchmark MuSiQue, the substantial gap likely stems from the significantly larger parameter count (235B vs 4B) and superior intrinsic reasoning capabilities of the baseline generator, whereas the performance differences on datasets like NarrativeQA and TriviaQA remain marginal.

\subsubsection{NIAH task}

\begin{figure}[h]
    \centering
    \includegraphics[width=1.0\linewidth]{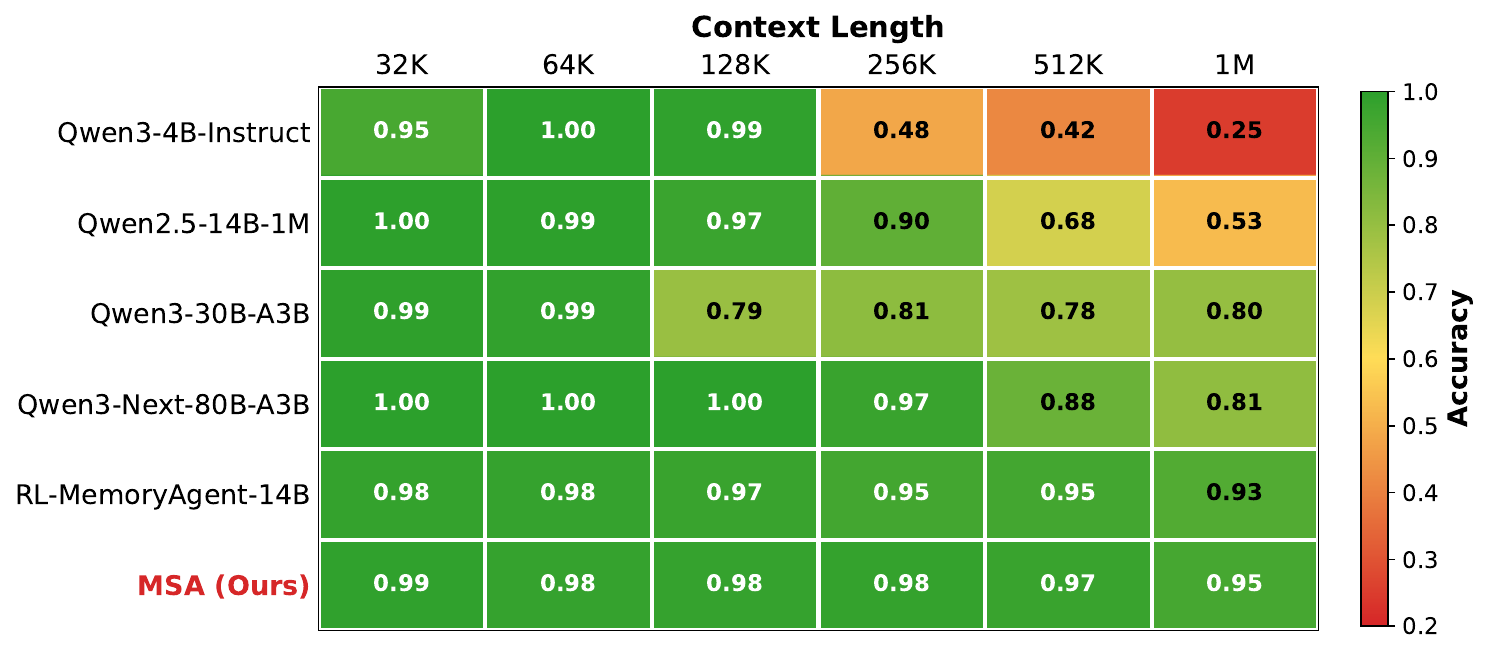}
    \caption{Results on the "Needle In A Haystack" (NIAH) evaluation across varying context lengths from 32k to 1M tokens.}
    \label{fig:niah_results}
\end{figure}

On the RULER Needle-In-A-Haystack (NIAH) benchmark, MSA demonstrates exceptional stability when scaling the context length from 32k to 1M tokens. As shown in Figure~\ref{fig:niah_results}, the model exhibits only a gradual accuracy decay across this 32-fold expansion, ultimately maintaining a high retrieval accuracy of 94.84\% at the 1M-token scale. In stark contrast, the unmodified backbone model (Qwen3-4B-Instruct) suffers catastrophic degradation beyond 128k tokens, with its accuracy plummeting to 48.16\% at 256k tokens and further deteriorating to 24.69\% at 1M tokens, rendering it practically ineffective for ultra-long contexts.

Hybrid linear attention models designed for long-context processing also exhibit significant instability under extreme scaling. Specifically, Qwen2.5-14B-1M experiences a sharp performance drop at 256k tokens (accuracy falling below 90\% to 89.97\%), Qwen3-30B-A3B shows severe degradation starting at 128k tokens (accuracy dropping to 79.13\%), and even the largest variant Qwen3-Next-80B-A3B, despite near-perfect performance up to 128k tokens, undergoes substantial decay beyond 256k tokens, with accuracy decreasing to 80.78\% at 1M tokens.

Among external storage-based memory approaches, RL-MemoryAgent-14B displays relatively stable performance without catastrophic failure points, yet its absolute accuracy remains consistently lower than MSA across all context lengths. More critically, its decay rate is markedly steeper: while MSA retains 94.84\% accuracy at 1M tokens, reflecting only a 3.93-percentage-point drop from its 32k-token performance of 98.77\%, MemoryAgent-14B declines to 92.66\% at the same scale, corresponding to a 5.76-percentage-point reduction from its 32k-token accuracy of 98.42\%. These results collectively validate that MSA's sparse attention mechanism with document-wise RoPE not only achieves superior absolute performance but also provides substantially enhanced robustness for extreme-long context extrapolation compared to both conventional long-context architectures and external memory systems.

\subsection{Ablation Study}
\label{sec:ablation}

\begin{table*}[!h]
\centering
\caption{Ablation study on four QA benchmarks. We compare the full MSA-S1 model against three variants: removing multi-round interleaved retrieval (\textit{w/o multi-round}), skipping continual pre-training with auxiliary routing supervision (\textit{w/o pretrain}), and disabling loading of original document text after document ID generation (\textit{w/o original text}). Additionally, we compare MSA-S2 and MSA-S1 to demonstrate the effect of curriculum learning. All scores are reported on a 0--5 quality scale.}
\label{tab:abl_results}
\resizebox{0.85\linewidth}{!}{
\begin{tabular}{l c c c c c}
\toprule
\textbf{Model Variant} & \textbf{Average} & \textbf{MS MARCO v1} & \textbf{Natural Questions} & \textbf{DuReader} & \textbf{HotpotQA} \\
\midrule
\textbf{MSA-S2 (Full)}            & \textbf{3.976} & \textbf{4.141} & \textbf{3.545} & \textbf{4.155} & \textbf{4.061} \\
\textbf{MSA-S1 (Full)}            & 3.694          & 3.197          & 3.493 & 4.064          & 4.020 \\
\hspace{8pt}\textbf{w/o memory interleave}   & 3.497          & 3.175          & 3.485          & 4.076 & 3.250 \\
\hspace{8pt}\textbf{w/o continual pre-training}     & 2.537          & 2.267          & 2.448          & 3.144          & 2.289 \\
\hspace{8pt}\textbf{w/o original text} & 2.325          & 2.625          & 2.190          & 2.186          & 2.297 \\
\bottomrule
\end{tabular}
}
\end{table*}

To systematically validate the contribution of each core component in MSA, we conduct comprehensive ablation experiments on four representative question answering benchmarks. As summarized in Table~\ref{tab:abl_results}, we evaluate four critical design choices: (1) the impact of the two-stage curriculum learning strategy (MSA-S2 vs. MSA-S1), (2) the memory interleave mechanism for multi-hop reasoning, (3) the continual pre-training stage, and (4) the integration of original document text after document ID generation.

\textbf{Impact of Curriculum Learning.} First, comparing the fully trained MSA-S2 against the first-stage MSA-S1 reveals a 7.6\% average performance gain conferred by the second-stage curriculum training that extends context length from 8k to 64k tokens. This improvement is especially pronounced on datasets with massive memory banks: on MS MARCO (7.34M tokens), MSA-S2 achieves a remarkable gain of 29.5\% over MSA-S1. These results provide compelling empirical validation that progressive exposure to longer contexts during training substantially enhances the model's ability to extrapolate to massive memory scales during inference.

\textbf{Impact of Memory Interleave.} Second, the memory interleave mechanism delivers substantial gains on complex reasoning tasks. Removing this capability from the MSA-S1 baseline results in a 5.3\% average performance degradation. The impact is particularly magnified on multi-hop datasets: HotpotQA experiences a significant 19.2\% drop. This pattern confirms that memory interleave, where the model refines its retrieval query based on previously acquired evidence, is essential for compositional reasoning that requires evidence chains.

\textbf{Impact of Continual Pre-training.} Third, CPT on large-scale retrieval tasks serves as the foundation for establishing robust routing capabilities. This stage incorporates a warmup phase that rapidly primes the router, significantly enhancing router-level precision, a factor pivotal to the model's overall retrieval efficacy. Eliminating CPT results in a severe average performance degradation of 31.3\% (dropping from 3.694 to 2.537). This deterioration is consistent across all benchmarks, with the multi-hop dataset HotpotQA suffering a massive 43.1\% decline. This sharp drop occurs because multi-hop tasks rely on memory interleaving, where errors in initial document retrieval accumulate during subsequent steps, thereby undermining the model's final reasoning performance.

\textbf{Impact of Original Text.} Fourth, integrating original document text after document ID generation provides substantial semantic grounding for answer synthesis. Disabling this component leads to the most severe average performance decline of 37.1\% (from 3.694 to 2.325). Tasks requiring detailed reading comprehension suffer immensely: DuReader exhibits a massive 46.2\% drop (4.064 to 2.186). This suggests that while document ID generation effectively localizes relevant evidence, the subsequent injection of raw document semantics remains essential for extracting the nuanced factual details necessary for precise response synthesis.

\section{Analysis}

In this section, we analyze the scalability of Memory Sparse Attention (MSA) focusing on two critical dimensions: computational efficiency and information fidelity. A robust long-memory model must satisfy dual constraints: ensuring linear computational complexity to make massive scaling feasible, and maintaining high generation quality (minimal context degradation) as the volume of noise increases. Our analysis confirms that MSA successfully bridges this gap, achieving $\mathcal{O}(L)$ efficiency while sustaining consistent QA performance with less than 9\% degradation across context lengths spanning from 16K to 100M tokens.

\subsection{Efficiency Analysis}
\label{sec:efficiency}

We analyze the computational complexity of Memory Sparse Attention with respect to memory size $L$. We define $M$ as the query length ($M \ll L$), $G$ as the average document length, $k$ as the number of top-$k$ documents selected (a small constant, e.g., 16), and $P$ as the chunk-wise pooling size (fixed at 64 in practice). MSA achieves linear complexity with respect to $L$ in both training and inference regimes, as detailed below.

\subsubsection{Training Complexity}

During training, MSA processes the entire memory bank within each forward pass. The computational cost consists of three components:

\begin{enumerate}
    \item \textbf{Independent Document Processing}: Each of the $L/G$ documents undergoes intra-document self-attention independently. With $\mathcal{O}(G^2)$ complexity per document, the aggregated cost across all documents is $\mathcal{O}(LG)$.
    
    \item \textbf{Sparse Routing}: The model computes relevance scores between the query representation and the $L/P$ pooled chunks from the memory bank, incurring $\mathcal{O}(ML/P)$ complexity.
    
    \item \textbf{Sparse Generation}: Attention is applied over the concatenated context comprising the query and the $k$ selected documents (each compressed to $G/P$ chunks). This stage costs $\mathcal{O}\big((M + kG/P)^2\big)$, which depends only on query length and fixed hyperparameters ($k$, $P$), and is therefore independent of the total memory size $L$.
\end{enumerate}

Summing these components yields the total training complexity:
\begin{equation}
\mathcal{O}_{\text{train}} = \mathcal{O}(LG) + \mathcal{O}(ML/P) + \mathcal{O}\big((M + kG/P)^2\big) = \mathcal{O}(LG)
\end{equation}
where the $\mathcal{O}(LG)$ term dominates when $L \gg G$, which holds in extreme-long memory scenarios.

\subsubsection{Inference Complexity}

MSA's inference pipeline separates computation into offline pre-processing and online query handling:

\begin{enumerate}
    \item \textbf{Offline Pre-processing}: Prior to serving queries, the system performs a one-time forward pass over the entire document collection to generate and cache compressed representations ($\bar{K}$, $\bar{V}$, $\bar{K}^R$) for all documents. This stage incurs $\mathcal{O}(LG)$ complexity but is executed only once per memory bank version—unlike conventional attention mechanisms that require $\mathcal{O}(L^2)$ prefill computation for every query.
    
    \item \textbf{Online Routing}: For each incoming query, the model matches the routing query against the precomputed cache of $L/P$ entries, costing $\mathcal{O}(ML/P)$.
    
    \item \textbf{Online Generation}: Autoregressive generation operates over the assembled sparse context of size $M + kG/P$. With $T$ denoting the answer length, this stage costs $\mathcal{O}\big(T \cdot (M + kG/P)^2\big)$, which remains independent of $L$.
\end{enumerate}

The per-query inference complexity is therefore:
\begin{equation}
\mathcal{O}_{\text{inference}} = \mathcal{O}(ML/P) + \mathcal{O}\big(T \cdot (M + kG/P)^2\big) = \mathcal{O}(L)
\end{equation}
where the routing term $\mathcal{O}(ML/P)$ dominates and scales linearly with memory size. Crucially, the expensive $\mathcal{O}(LG)$ pre-processing is amortized across all queries served from the same memory bank, yielding substantial efficiency gains in query-heavy workloads compared to methods requiring per-query $\mathcal{O}(L^2)$ prefill operations.

\subsection{Context Degradation}
\label{sec:context_degradation}

A persistent challenge in scaling language models is context degradation, where the accumulation of massive irrelevant context dilutes the model's ability to reason effectively. We evaluate MSA's robustness against this phenomenon using the MS MARCO Question Answering benchmark, measuring the QA score via an LLM judge as the memory context extends from 16K up to an unprecedented 100 million tokens. As illustrated in Figure~\ref{fig:scaling}, while state-of-the-art long-context models such as GPT-4.1 and DeepSeek-V3.2~\cite{liu2025deepseek} begin with competitive scores (approximately 3.6--3.7), they exhibit visible performance declines as the context length increases. In contrast, MSA demonstrates exceptional stability, starting with a strong score of 4.023 at 16K tokens and sustaining a competitive 3.669 even at the extreme 100M token scale. This represents a gradual degradation of only 8.8\% across four orders of magnitude in memory scaling. Conversely, the standard Qwen3-4B-Instruct backbone begins to degrade severely around 128k tokens and suffers a catastrophic collapse by 512k tokens (score dropping below 1.5). These results empirically validate that our sparse routing and independent positional encoding mechanisms effectively decouple reasoning capabilities from memory capacity, enabling MSA to operate reliably on massive-scale knowledge bases.
% \newpage
\section{Conclusion}
We introduce MSA, a scalable sparse‑attention framework augmented with document‑wise RoPE and KV‑cache compression that extends end‑to‑end modeling to lifetime‑scale contexts,
paired with Memory Parallel for fast 100M tokens processing and Memory Interleave for robust multi‑hop reasoning across distributed memory segments.
On long‑context QA and Needle‑in‑a‑Haystack benchmarks, MSA surpasses mainstream state‑of‑the‑art general‑purpose LLMs while preserving retrieval fidelity and reasoning depth, with KV‑cache compression further reducing memory footprint and latency.
Crucially, performance degradation remains minimal even under extreme context lengths, maintaining high accuracy as the effective context scales to 100M tokens.
These results indicate that by effectively decoupling memory capacity from reasoning capabilities, MSA can serve as a new foundational component to empower general‑purpose models with memory capacity.

\section{Limitations}
Although this work enhances intrinsic latent‑state memory for long textual contexts,
it remains limited when tasks require modeling strong and tightly coupled dependencies across multiple documents.
In scenarios where evidence is distributed and highly interlinked across sources,
the method struggles to maintain accurate structural alignment purely through intrinsic memory.
Memory interleave is a potentially promising direction for mitigating these issues,
as it can help integrate and synchronize information from separated context segments.
However, its effectiveness depends on more efficient and principled designs that better preserve inter‑document relationships.
\bibliography{custom}

\newpage
\appendix

\section{Prompts}
\label{sec:appendixA}

\NewTColorBox{EqBox}{ s O {!htbp} m }{%
  floatplacement={#2},
  IfBooleanTF={#1}{float*,width=\textwidth}{float},
  title={\textsc{#3}},
}

\begin{EqBox}[!htbp]{Prompt Template for LLM as a Judge}
\vspace{1mm}
{\tt \footnotesize  
Based on the accuracy, completeness, and relevance of the predicted answer to the real answer in the context of the **query**, assign an objective score from 0 to 5 (5 being the highest, 0 the lowest).
\\
\\
    The scoring must strictly adhere to the following criteria. The final output can only be a single number.
\\
\\
    Scoring Criteria:
\\
\\
    5: The predicted answer is exactly the same as the real answer and correctly answers the query. Differences in wording do not affect factual accuracy.
\\
\\
    4: The predicted answer contains all the core information of the real answer, with no errors, but includes a small amount of non-critical redundant content.
\\
\\
    3: The predicted answer captures the core information but differs from the real answer in some aspects. The predicted answer is slightly incomplete or imprecise, but contains no errors.
\\
\\
    2: The predicted answer is partially relevant to the real answer but omits a significant amount of information or deviates from the core topic of the query.
\\
\\
    1: The predicted answer attempts to address the query (maintains basic relevance to the topic) but provides factually incorrect information. It does not contradict the core claim of the real answer, but shows incomplete or inaccurate understanding of the topic.
\\
\\
    0. The predicted answer is completely unrelated to the query, consists of gibberish, or is a pure hallucination that shares no logical connection with the real answer.
\\
\\
    Query:
\\
\\
    \{query\}
\\
\\
    True Answer:
\\
\\
    \{gold\_answer\}
\\
\\
    Predicted Answer:
\\
\\
    \{model\_answer\}
\\
\\
    Output only a single number (0, 1, 2, 3, 4, or 5): 
}
\end{EqBox}

\section{Pre-training Data Composition}
\label{app:pretrain_data}

\begin{table*}[h]
\centering
\caption{Detailed statistics of the full MSA Pre-training Dataset.}
\label{tab:pretrain_data}
\resizebox{0.9\linewidth}{!}{
\begin{tabular}{l r r l}
\toprule
\textbf{Dataset Source (Filename)} & \textbf{Queries} & \textbf{Tokens} & \textbf{Task / Domain} \\
\midrule
\multicolumn{4}{l}{\textit{\textbf{Long-Context \& Instruction Tuning}}} \\
\texttt{kalmfinetune\_data} & 5,801,540 & 6.46B & Knowledge Augmentation \\
% \texttt{kalm\_funetune\_5pos} & 332,851 & 5.03B & Long-Context Retrieval \\
% \texttt{kalm\_funetune\_5posv2} & 554,755 & 8.39B & Long-Context Retrieval \\
\midrule
\multicolumn{4}{l}{\textit{\textbf{Academic \& Scientific Literature}}} \\
\texttt{S2ORC\_citations\_abstracts} & 500,000 & 7.31B & Scientific Literature \\
\texttt{S2ORC\_citations\_titles} & 500,000 & 7.18B & Scientific Literature \\
\texttt{S2ORC\_title\_abstract} & 500,000 & 7.11B & Scientific Literature \\
\texttt{specter\_train\_triples} & 500,000 & 7.14B & Scientific Citation \\
\midrule
\multicolumn{4}{l}{\textit{\textbf{General QA \& Community Knowledge}}} \\
\texttt{yahoo\_answers\_qa} & 500,000 & 7.10B & General QA \\
\texttt{yahoo\_answers\_ta} & 500,000 & 7.09B & General QA \\
\texttt{yahoo\_answers\_tq} & 500,000 & 7.10B & General QA \\
\texttt{WikiAnswers} & 500,000 & 7.16B & Community QA \\
\texttt{gooaq\_pairs} & 500,000 & 7.11B & FAQ / Common QA \\
\texttt{msmarco\_triples} & 499,184 & 7.37B & Information Retrieval \\
\texttt{PAQ\_pairs} & 500,000 & 7.15B & Synthetic QA \\
\texttt{amazon\_qa} & 500,000 & 7.11B & E-commerce QA \\
\texttt{eli5\_question\_answer} & 325,475 & 4.62B & Explainable QA \\
\texttt{stackexchange\_body\_body} & 250,460 & 3.58B & Technical QA \\
\texttt{stackexchange\_title\_body} & 250,519 & 3.59B & Technical QA \\
\texttt{stackexchange\_title\_title} & 304,525 & 4.31B & Technical QA \\
\texttt{searchQA\_top5\_snippets} & 117,220 & 1.68B & Machine Reading Comprehension \\
\texttt{quora\_duplicates} & 103,663 & 1.46B & Duplicate Detection \\
\texttt{quora\_duplicates\_triplets} & 101,762 & 1.45B & Duplicate Detection \\
\texttt{NQ\_train\_pairs} & 100,231 & 1.43B & Open-Domain QA \\
\texttt{squad\_pairs} & 87,599 & 1.24B & Reading Comprehension \\
\texttt{TriviaQA\_pairs} & 73,346 & 1.04B & Trivia QA \\
\midrule
\multicolumn{4}{l}{\textit{\textbf{News \& Summarization}}} \\
\texttt{agnews} & 500,000 & 7.10B & News Classification \\
\texttt{npr} & 500,000 & 7.01B & News Broadcast \\
\texttt{ccnews\_title\_text} & 500,000 & 6.96B & News Data \\
\texttt{cnn\_dailymail\_splited} & 311,971 & 4.70B & Long-Document Summarization \\
\texttt{cnn\_dailymail} & 311,971 & 4.41B & News Summarization \\
\texttt{xsum} & 226,711 & 3.20B & Extreme Summarization \\
\texttt{sentence\_compression} & 180,000 & 2.55B & Text Compression \\
\texttt{altlex} & 112,696 & 1.60B & Paraphrasing \\
\midrule
\multicolumn{4}{l}{\textit{\textbf{Domain-Specific \& Others}}} \\
\texttt{amazon\_review\_2018} & 500,000 & 7.08B & E-commerce Reviews \\
\texttt{codesearchnet} & 500,000 & 7.01B & Code / Programming \\
\texttt{AIINLI} & 277,230 & 3.94B & Natural Language Inference \\
\texttt{wikihow} & 128,543 & 1.82B & Instructional / How-to \\
\texttt{SimpleWiki} & 102,225 & 1.46B & General Knowledge \\
\texttt{coco\_captions} & 82,783 & 1.18B & Image Captioning \\
\texttt{flickr30k\_captions} & 31,783 & 0.45B & Image Captioning \\
\midrule
\textbf{Total} & \textbf{17,852,825} & \textbf{158.95B} & \textbf{All Domains} \\
\bottomrule
\end{tabular}
}
\end{table*}

To ensure the model possesses both robust retrieval capabilities and broad general knowledge, we constructed a diverse pre-training corpus comprising 158.95 billion tokens across 17.9 million queries. As detailed in Table~\ref{tab:pretrain_data}, the corpus covers a wide range of domains from scientific literature to general community Q\&A. To maintain a balanced data distribution, we downsample any dataset outside the KALM suite that exceeds 0.5 million queries to a maximum of 0.5 million, while retaining the KALM instruction data in its entirety.

% \section{Case Study}

\end{document}